\documentclass[review]{elsarticle}
\usepackage[OT1]{fontenc}
\usepackage{lineno,hyperref}

\usepackage{times}
\usepackage{epsfig}
\usepackage{graphicx}
\usepackage{amsmath}
\usepackage{amssymb}
\usepackage{booktabs}
\usepackage{bm}
\usepackage{setspace}
\usepackage[medium,compact]{titlesec}
\usepackage{enumitem}
\usepackage{array}

\usepackage{multirow}

\journal{Journal of Expert Systems with Applications}

\begin{document}

\begin{frontmatter}

  \title{Geometry Attention Transformer with Position-aware LSTMs for Image Captioning$^2$}

  %% Group authors per affiliation:
  % \author{Elsevier\fnref{myfootnote}}
  % \address{Radarweg 29, Amsterdam}
  % \fntext[myfootnote]{Since 1880.}

  %% or include affiliations in footnotes:
  % \author{Chi Wang\fnref{firstauthor}}
  % \ead{chi.w@std.uestc.edu.cn}

  \author{Chi Wang\fnref{firstauthor}}
  \ead{chi.w@std.uestc.edu.cn}
  \author{Yulin Shen\fnref{firstauthor}}
  \fntext[firstauthor]{These authors contributed equally to this work.}
  \ead{yulinshen@std.uestc.edu.cn}

  \author{Luping Ji\corref{mycorrespondingauthor}}
  \cortext[mycorrespondingauthor]{Corresponding author.}
  \ead{jiluping@uestc.edu.cn}

  \address{School of Computer Science and Engineering, University of Electronic Science and Technology of China, \\ Chengdu 611731, P.R. China}

  \begin{abstract}
In recent years, transformer structures have been widely applied in image captioning with impressive performance. For good captioning results, the geometry and position relations of different visual objects are often thought of as crucial information. Aiming to further promote image captioning by transformers, this paper proposes an improved \textit{Geometry Attention Transformer} (GAT) model. In order to obtain geometric representation ability, two novel geometry-aware architectures are designed respectively for the encoder and decoder in our GAT. Besides, this model includes the two work modules: \romannumeral1) a geometry gate-controlled self-attention refiner, for explicitly incorporating relative spatial information into image region representations in encoding steps, and \romannumeral2) a group of position-LSTMs, for precisely informing the decoder of relative word position in generating caption texts. The experiment comparisons on the datasets MS COCO and Flickr30K show that our GAT is efficient, and it could often outperform current state-of-the-art image captioning models.

  \end{abstract}

  \begin{keyword}
    Image captioning, Encoder-decoder framework, Gate-controlled geometry attention, Position-aware LSTM
  \end{keyword}

  \fntext[]{The source code is available at \url{https://github.com/UESTC-nnLab/GAT}}

\end{frontmatter}

% \linenumbers

\section{Introduction}
%先介绍image captioning的总体情况。
Image Captioning is a challenging problem in computer vision\cite{farhadi2010every}. It aims to automatically describe an image using meaningful text, translating an image into natural language. It often requires to not only recognize what visual objects an image contains but also capture what those objects are doing accurately and even to tell us what the interrelations of different objects are. Image captioning could build a powerful bridge between visual images and human languages. With captions, people could better understand an image\cite{krishna2017visual}. Image captioning has been thought of as one quite useful technology in the image analysis field. In recent years, it has been attracting more and more research attention, and diverse captioning models have been emerging for image captioning \cite{rennie2017self, Anderson_2018_CVPR, cornia2019m, Zhangjing2021}.

%介绍image captioning的相关算法模型
For image captioning, its early models could often be roughly classified into two primary categories \cite{Baishuang2018}. One is usually dependent on image retrieval. It generates image captions by analyzing image correlation and then retrieving candidate texts from existing caption pools \cite{Ordonez2011nips, Gupta2012}. In contrast, the other is often summarized as the template-based category. This kind of methods builds captions typically through the syntactic and semantics analysis on images, with visual concept detecting, sentence template matching and optimizing \cite{socher2010connecting, Kulkarni2013, Ushiku2015}. The most obvious characteristic of these two early categories is that hardcoded rules and hand-engineered features are in common use so that the reliability and accuracy of captioning fluctuate heavily \cite{Baishuang2018}.

%话题过渡到本文的重点，深度神经网络
In recent years, facilitated by booming artificial intelligence, neural networks, the third type of method for image captioning, have been becoming one of the most exciting and powerful tools. For example, we could easily see the huge improvements by early work of deep neural models in \cite{Karpathy2014, Malin2015, Yanfei2015}. Besides, more neural network based models, such as the multimodal learning \cite{karpathy2015deep,karpathy2017deep}, the encoder$-$decoder framework \cite{vinyals2015show}, the attention mechanism \cite{Youquanzeng2016,huang2019attention}, the compositional architectures \cite{Oruganti2016}, the describing method of novel objects \cite{Maojunhua2015} and the deep bifurcation network \cite{NABATI2021115541}, have been proposed one after another.

Through the survey of a large number of publications, it is easy to figure out that Transformer with an encoder-decoder structure is becoming one of the most mainstream models for image captioning \cite{guo2020normalized}. In general, in a typical transformer model, a given image is often mapped into a set of intermediate vectors by an encoder of CNN-based networks, and then the target caption of this image could be generated word by word through a decoder of RNN-based network \cite{Anderson_2018_CVPR,karpathy2015deep,vinyals2015show,Lu_2017_CVPR}.

%
%\begin{figure}[htb]
%    \centering
%    \includegraphics[width=0.5\textwidth]{figures/model.png}
%    %\includegraphics[scale=0.6]{figures/model.png}
%    \caption{Overview of the proposed captioning model. The geometry self-attention layer in the encoder refines appearance features with geometry features. The decoder generates the caption using the refined appearance features and the sequence position context.}
%    \label{model}
%\end{figure}

In addition, as mentioned before, the classic attention mechanism has also been widely utilized in image captioning \cite{Youquanzeng2016,huang2019attention}. One of its advantages is that it could often effectively guide the decoder to focus on some specific information, such as context correlation, while generating image captions. On traditional attention models, the self-attention mechanism \cite{vaswani2017attention} is often thought of as an effective improvement to the classic version. In recent years, this kind of self-attention model has also quickly emerged in image captioning \cite{herdade2019image,li2019entangled}. From the task perspective, a self-attention module is often regarded as a feature mapper between a set of queries and key/value pairs.

In image captioning, the geometry and position relations of image objects are often very necessary for accurately describing a given image. For example, ''\emph{a boy standing on a skateboard}'' and ''\emph{a boy raises a skateboard in his hands}'' indicate two different semantics. However, they are likely to mean the same thing if without relative geometry information. In most existing Transformer-based models with attention modules, we could find that the inherent geometry and position relations between image components have not yet been paid enough attention to and fully utilized \cite{Lu_2017_CVPR}. Moreover, although some methods, such as the GSA proposed in \cite{guo2020normalized}, carefully considers geometry relations in encoder, more detailed position decoding has not been well solved yet.

%
%As plausible as it sounds, however, feeding the vanilla self-attention module with raw object vectors fails to model the geometrical relationships among the detected objects. The geometrical structure, such as relative positions and sizes, is essential to reasoning in reality. Admittedly, ``a boy standing on a skateboard" and ``a boy raises a skateboard in his hands'' may ridiculously mean the same thing without the relative geometrical information.

Geometry and position relations are so important that accurate image captioning models often have to emphasize them. In order to take full advantage of geometry and position clues, we firstly design an improved encoder model with {Geometry self-attention Refiner} (GSR) in our transformer structure. This model could explicitly incorporates geometry information into a vanilla self-attention module, implementing the transformation from original appearance queries to geometrical ones. And then, the refined attention weights are calculated to get a mean of weighted appearance values. The geometrical queries and keys in our model are linearly derived from the bounding box coordinates of each component object. In this way, every object could always obtain its appearance features and geometry correlations with each other. In task details, they both are very crucial for a decoder to generate the right caption word sequences with position and semantic.

%Furthermore, in the the decoder of transformer architecture, due to the born nature of simply treating sequence as ``bag-of-words'', the self-attention module cannot inevitably model the order of the words in the input sequence. Notably, the position and the order of words define the grammar and thus the actual semantics of a sentence, which one cannot afford to lose. The Transformer architecture abandons the recurrence mechanism and suffers from representing the sequential feature of a sentence. Prior works typically inject position encodings by adding sines and cosines on top of the actual word embeddings\cite{vaswani2017attention, herdade2019image}. However, this mechanism can be semantically insufficient and thus restrict the capacity of the subsequent modules.

Furthermore, in the decoder of Transformer, some typical methods often inject the word position information of expected caption texts by adding \emph{sine} or \emph{cosine} operation on the top of the word embedding layer, such as the ones in \cite{vaswani2017attention, herdade2019image}. This kind of position injection mechanism has been proved effective, in spite of not always conforming to semantics. To overcome this weakness of position decoding, we design a group of position-LSTMs to model the word order of caption texts. It parses a caption sentence word by word in sequence, ensuring the right order of words. In addition, the hidden layers of position-LSTMs could also remember the information that the decoder has generated. This kind of design is very helpful to self-attention modules to concentrate on particular position parts. In the meanwhile, the information stored in hidden states of LSTMs also contains the geometry information of component objects.

Finally, by the integration and collaborative optimization of {GSR} and the {position-LSTM}, our captioning model has achieved better performance than state-of-the-art ones on the two classic datasets: MS COCO and Flickr30k.
%Figure \ref{model} depicts the overall architecture of our proposed GAT model.
In summery, the main contributions of this paper could be listed as follows:
% The main contributions of this paper are as follows:
\begin{list}{\labelitemi}{\leftmargin=1em}
    \setlength{\topmargin}{0pt}
    \setlength{\itemsep}{0em}
    \setlength{\parskip}{0pt}
    \setlength{\parsep}{0pt}

    % \vspace{-0.3cm}
    \item We propose an improved image captioning model {GAT} of transformer structure, which could capture the geometry relations of component objects. Geometry capturing ability could often enable our captioning model to obtain a sense of \emph{'where are target objects'} and \emph{'where is the captioning model currently looking at'}.

    \item We design an encoder improved by {GSR}. It could encapsulate the relative geometry information of objects. Moreover, it takes full advantage of geometry relations to refine object representations.

    \item We develop a decoder promoted by position-LSTMs. It could transport the order information of generated cation words with relative position relations. Therefore, it could execute position reasoning to decide which word to generate next. %This manner, to our knowledge, provides the community with a new direction to explore the position encodings of Transformer-like models.
    % \item By combining the \textit{GSR} and the \textit{position-LSTM}, our model achieves the state-of-the-art performance on the COCO dataset.
\end{list}
% \vspace{-0.4cm}
\section{Related Work}

 In recent years, a king of encoder-decoder transformer architecture with attention modules has been attracting wide research enthusiasm. In the meanwhile, more and more methods have been developed for image captioning, such as the one with soft $\&$ hard attention \cite{xu2015show}, the one with residual connections \cite{gao2019deliberate}, and the meshed-memory transformer \cite{ cornia2020meshed}.

\begin{figure}[http]
  \small
  \centering
  \resizebox{\linewidth}{!}{
  \begin{tabular}{cc}
    \includegraphics[width=0.6\textwidth]{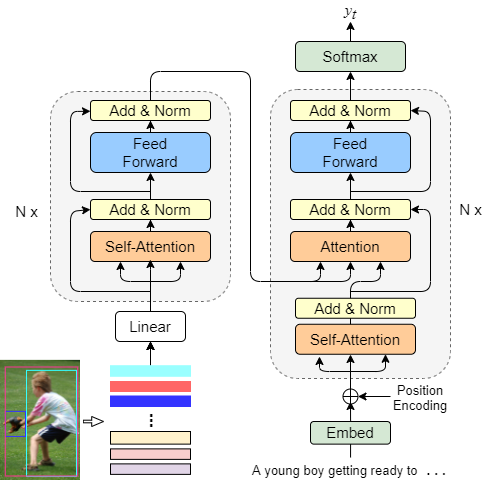} &
    \includegraphics[width=0.25\textwidth]{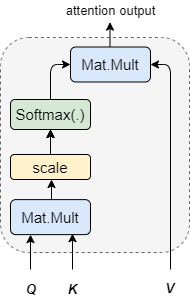} \\
    \addlinespace
    (a) & (b)
  \end{tabular}}
  \caption{a typical deep neural network of image captioning \cite{guo2020normalized}, (a) encoder-decoder structure, (b) a classic self-attention component widely integrated in both encoder and decoder.
  }

 % The architecture of the self-attention network. This image captioning model leverages a vanilla Transformer and could be regarded as a baseline. \quad  b) Overview of our proposed captioning model. The geometry self-attention layer in the encoder refines appearance features with geometry features. The decoder generates the caption using the refined appearance features and the sequence position context.}
  \label{fig:transformer}
\end{figure}

 %However, none of these methods take the relative relationships among objects into account.
This type of transformer architecture, shown in Figure \ref{fig:transformer}, can bee clearly seen in \cite{guo2020normalized,herdade2019image,zhou2020unified,Chenhaishun2021}. Typically, it consists of an encoder and a decoder, one for extracting visual object features and the other for generating caption text. In detail, both the encoder and the decoder, represented by Fig. \ref{fig:transformer}(a), are designed as a group of multi-layer residual networks. Moreover, we could also always see a classic component of '\textbf{Self-Attention}', shown by Fig. \ref{fig:transformer}(b), in the encoder and decoder \cite{herdade2019image,li2019entangled}.

In the typical captioning framework illustrated above, besides traditional object appearance features, the input to the encoder contains some novel geometry cues, such as object center, height and width \cite{guo2020normalized}. In the decoder, furthermore,  a position encoding operation has also been employed to encode the accurate word sequence of the caption text. And in \cite{guo2020normalized}, a simple 'sinusoidal function' is utilized to fulfill this task. Experiments have also proved that this kind of improvement by geometry and position features is much efficient.

Moreover, in terms of the attention modules in image captioning, there are also all kinds of variants having been explored. For examples, we could easily see the multi-head attention in \cite{vaswani2017attention}, the gate-controlled attention \cite{Oruganti2016}, the fully attentive paradigms \cite{li2019entangled,zhu2018captioning}, the meshed-connection attention \cite{cornia2020meshed}, the dual attention \cite{Yulitao2021}, etc.

To our knowledge, this kind of encoder-decoder architecture with attention components introduced above is believed to be one of the most potential baselines for image captioning \cite{guo2020normalized,herdade2019image}. In the future, more ingenious network design, more sophisticated object features,  more functional geometry relations, and position encoding strategies are still kept huge research expectations for further promoting image captioning methodology \cite{Liushen2021Sibnet}.

%However, the geometrical relationships among the detected objects are still ignored.

% 比较相似模型之间的差异
%The most similar works to our work are the ones of Herdade et al.\cite{herdade2019image} and Guo et al.\cite{guo2020normalized}. Herdade et al. use an object relation module to model the geometry information. Guo et al. incorporate a naive Transformer-like architecture, referred to as SAN, into the image captioning task, and further calculate a geometric bias to modify the content-based weight to get an improved model, GSA. The architecture of SAN is demonstrated in Figure \ref{fig:model} and could be regarded as a baseline to our work. Our approach employs a simple but efficient way to further adaptively model the inherent geometrical relationships. Meanwhile, we note that the positional encoding in the decoder remains to be exploited.

\section{Proposed Method}
\begin{figure}[htb]
  \begin{center}
    \includegraphics[width=.98\textwidth]{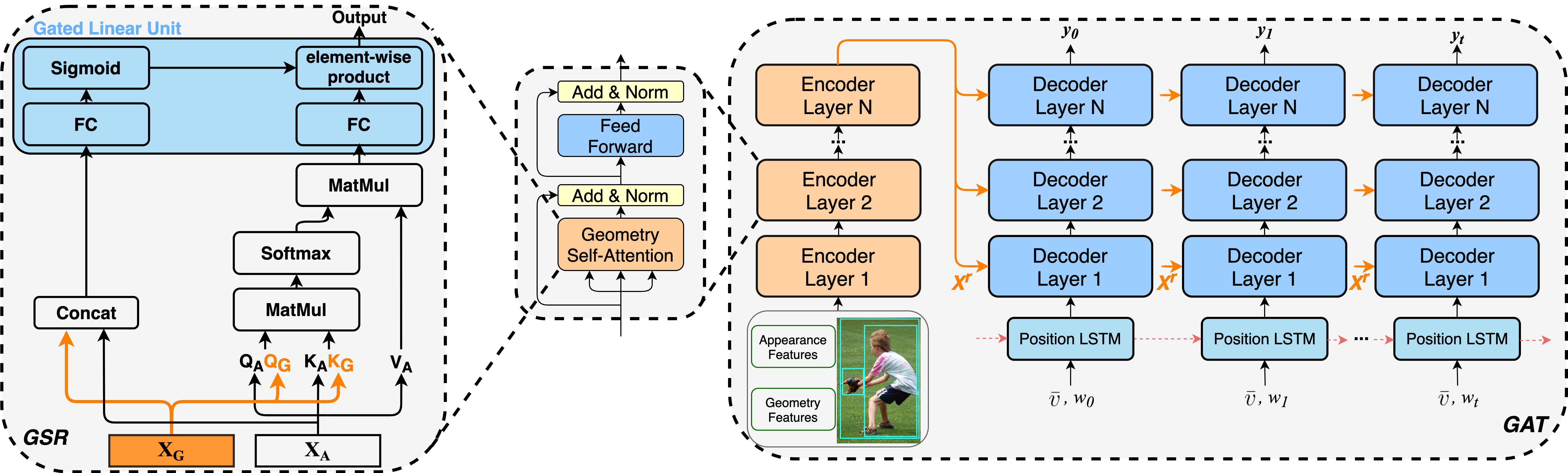}
  \end{center}
  \caption{The detailed architecture of our GAT. The GSR module in encoder generates some weighted average given appearance features \(\bm{X}_A\) and geometry features \(\bm{X}_G\), which is further refined by a gate-controlled unit. Moreover, a group of position-LSTMs precisely injects word order information into the decoder.}
  \label{fig:enc-dec}
\end{figure}

Our method generates grounded captions by attending to specific image regions at each step. In terms of structure, it still retains a classic encoder-decoder architecture. In the encoder, a geometry self-attention refiner (GSR) is designed to optimize image representations. Furthermore, in the decoder, a module for accurately decoding word sequences by LSTM is adopted. More details about it can be seen in Figure \ref{fig:enc-dec}.

Given an input image \textbf{\emph{I}}, in figure \ref{fig:enc-dec}, we use $(\bm{X}_A \in \mathbb{R}^{N \times d})$ to represent a set of image appearance regions, where $N$ is the index of image region, and $d$ is the dimensionality of region data. Moreover, $(\bm{y} = \{\bm{y}_1, ..., \bm{y}_T\})$ to indicate the caption word vectors corresponding to \textbf{\textit{I}}.

\subsection{Gate-controlled Geometry Self-attention Refiner}
% Generally, our geometrical self-attention refiner sub-model incorporates spatial relationships into self-attention networks and refines appearance features in a multi-layer fashion.
% In this section, we first briefly review the basic form of self-attention network and then describe our proposed geometrical self-attention refiner in detail.

% \textbf{Vanilla self-attention.}
% A basic self-attention module first transforms an input matrix \(X \in \mathbb{R}^{N \times d_{k}}\) into queries \(Q\), keys \(K\) and values \(V\) given by \(Q=XW_Q, K=XW_K, V=XW_V\) respectively, where \(W_Q, W_K, W_V\) are all learnable weights of dimension \(d_k \times d_{model}\). Then it measures the similarities \(\Omega\) between any queries and keys by:
% \begin{equation}
%     \Omega = \frac{QK^{T}}{\sqrt{d_k}}\label{att1}
% \end{equation}
% where \(\Omega\) is an \(N \times N\) attention weights matrix with which each element $\omega^{mn} $ representing the attention weights between the m-th and n-th token. The next step is to compute a weighted average of the values. In a scaled dot-product manner, the above process can be formulated as:
% \begin{equation}
%   % f_{self-att}(X) = {\rm softmax}(\Omega)V \label{self-att}
%   {\rm Attention}(X) = {\rm softmax}(\Omega)V \label{self-att}
% \end{equation}

% \textbf{Geometrical self-attention refiner.}
Besides appearance information \(\bm{X}_A\), we incorporate geometry information into the vanilla self-attention to refine the representation of the objects for the reason that it is beneficial to comprehend the intrinsic relations among different objects. Therefore, we propose the Geometry Self-attention Refiner (GSR) to refine the visual information by taking into account the geometry information of the objects.

Given the geometry features of the objects \(\bm{X}_g \in \mathbb{R}^{N \times 5}\), each row of \(\bm{X}_g\) is a 5-dimensional vector:
% \vspace{-0.2cm}
\begin{equation}
  (x_{min}, y_{min}, x_{max}, y_{max}, S)
  % % \vspace{-0.2cm}
\end{equation}
where \((x_{min}, y_{min})\), \((x_{max}, y_{max})\) and \(S\) represent the top-left coordinates, the bottom-right coordinates of bounding box, and the relative size to whole image, respectively. And, all of them  are normalized to $(0, 1)$.

Firstly, we obtain \(\bm{X}_{G}\in \mathbb{R}^{N \times d_m}\) by embedding \(\bm{X}_{g}\) to a higher dimension form with an embedding layer followed by a ReLU non-linearity. Then we combine appearance information with geometry information by modifying the queries and keys as follows:
% \vspace*{-0.85\baselineskip}
\begin{align}
  \bm{Q^{'}} & = [\bm{X}_A\bm{W}_{\bm{Q}_A}; \bm{X}_G\bm{W}_{\bm{Q}_G}] \\
  \bm{K^{'}} & = [\bm{X}_A\bm{W}_{\bm{K}_A}; \bm{X}_G\bm{W}_{\bm{K}_G}]
  % \vspace*{-0.25\baselineskip}
\end{align}
where \(\bm{W}_{\bm{Q}_A}\), \(\bm{W}_{\bm{K}_A}\) are learned appearance matrices and  \(\bm{W}_{\bm{Q}_G}\), \(\bm{W}_{\bm{K}_G}\) are learned geometry matrices, respectively, and are all of dimension \(\mathbb{R}^{d_{m} \times d_{m}}\). Here [;]  indicates concatenation. \(\bm{Q^{'}}\) and \(\bm{K^{'}}\) combine appearance information with geometrical information, which can be seen as a complementary to acquire  fine-grained knowledge of the objects. Then the attention can be calculated as:
% \vspace*{-0.6\baselineskip}
\begin{equation}
  \bm{\Omega^{'}} = \frac{\bm{Q^{'}K^{'T}}}{\sqrt{2 \times d_k}}
  % \vspace*{-0.15\baselineskip}
\end{equation}
Then the output of the GSR can be calculated as:
% \vspace*{-0.5\baselineskip}
\begin{equation}
  { Attention_{g}}(\bm{X}) = { softmax} (\bm{\Omega^{'}})\bm{V_A}
  % \vspace*{-0.5\baselineskip}
\end{equation}

Similar to the original Transformer\cite{vaswani2017attention}, our geometry refiner is also implemented in a multi-head self-attention framework. In this framework, geometry attention will be repeated \(h\) times (so, called \(h\) heads). In every time of repetition, it always adopts a group of different projection matrices $(\bm{W}_{\bm{Q}_{A}}^{k}, \bm{W}_{\bm{K}_{A}}^{k}$ and $\bm{W}_{\bm{V}_{A}}^k)$ for extracting appearance features, and $(\bm{W}_{\bm{Q}_{G}}^{k}$ and $\bm{W}_{\bm{K}_{G}}^{k})$ for capturing geometrical features, where \(k\in(1,2,...,h)\). All the results generated from \(h\) self-attention heads will also be concatenated together by a linear projection and then fed forward to the next layer.

Moreover, inspired by the work of Huang et al. \cite{huang2019attention}, we further design a parallel Gate-controlled Linear Unit (GLU) to continually refine the output of the geometry self-attention module (the left region of Figure.\ref{fig:enc-dec}). This GLU still takes the current context $(\bm{\tilde{c}} = [\bm{X}_A; \bm{X}_G])$ as its input to generate a control matrix \(\bm{Gate_{Ctrl}}\) for modulating final attention output by Hadamard product. This group of neural computation can be mathematically expressed by Eqn. (\ref{gate_contrl}):

% multili an information vector \(I\) by two separate linear transformations, respectively. The gate \(g\) is then applied to \(i\) to obtain final attention \(\hat{i}\):
% \begin{spacing}{1}
% \vspace*{-0.4\baselineskip}
\begin{equation}
\left\{
\begin{aligned}
\bm{Gate_{Ctrl}} &=sigmoid(\bm{W}_{g}\bm{\tilde{c}} + \bm{b}_g) \\
\bm{Output} &= \bm{Gate_{Ctrl}} \odot (\bm{W}_{i}\bm{\tilde{a}} + \textbf{b}_{i})
\end{aligned} \right.
\label{gate_contrl}
\end{equation}
where \(\bm{W}_{g}, \bm{W}_{i}\in \mathbb{R}^{d_{m} \times d_{m}}\) are two groups of weights, and \(\bm{b}_g, \bm{b}_{i} \in \mathbb{R}^{d_{m}}\) are two groups of neuron biases. Furthermore, \(\odot\) denotes a Hadamard product. In terms of structure, our self-attention component and gate-controlled unit can be stacked into multiple layers to further optimize object representations.
% \end{spacing}

Subsequently, the self-attention output is  embedded with its original input by the operation of add \& Normalization, then transferred to its next layer, $i.e.$, the \emph{Feed-Forward} sub-lay network shown in the middle of Figure \ref{fig:enc-dec}. In details, this \emph{Feed-Forward} sub-layer usually contains two groups of nested affine transformations with the activation function, $ReLU(x)=max(0,x)$. Therefore, this group of processing step could be uniformly formulated by
% \vspace*{-0.4\baselineskip}
\begin{equation}
   FF(\bm{X}) = { max}(0, \bm{X}\bm{{W}}_1 + \bm{b}_1)\bm{W}_2 + \bm{b}_2 \label{ff}
  % \vspace*{-0.4\baselineskip}
\end{equation}
where \(\bm{W}_1\) and \(\bm{W}_2\) are two groups of weight matrices to be learned, $\bm{b}_1$ and $\bm{b}_2$ are two groups of biases. Also following the \emph{Transformer} architecture in \cite{vaswani2017attention}, each sub-layer is successively followed by a residual connection and a layer normalization. Therefore, this group of computation could be mathematically expressed by
\begin{equation}
\left\{
\begin{aligned}
\bm{Z} &=LayerNorm(\bm{X} + Attention_g(\bm{X})) \\
\bm{X}^r &= LayerNorm(\bm{Z} + FF(\bm{Z}))
\end{aligned} \right.
% \label{gate_contrl}
\end{equation}

%% \vspace{-0.3cm}
%\begin{align}
%  Z   & = {\rm LayerNorm}(X + {\rm Attention_g} (X)) \\
%  % \vspace{-0.1cm}
%  X^r & = {\rm LayerNorm}(Z + {\rm FF}(Z))
%  % \vspace{-0.1cm}
%\end{align}

\subsection{Position-aware Self-attention Decoder}
% Our decoder is both conditioned on the previously generated words and the refined region features.
% In this section, we first describe the position-LSTM and then describe the self-attention-based position-aware caption decoder.

\textbf{i) Encoding positions via LSTM.}

In order to address the order problem of a sequence, existing works usually inject some '\emph{positional encoding}' into word embedding by a sinuous function. In our method, we further exploit a position representation strategy and then devise a distinctive LSTM-based position encoding mechanism.

The position-LSTM in our decoder of transformer could model the order of image caption words in decoding process. At each time step \(t\), we define the input of the position-LSTM as:
% \vspace*{-0.4\baselineskip}
\begin{equation}
  \bm{x}_{t} = [\bm{w}_t, \bm{\bar{v}}]
  % \vspace*{-0.4\baselineskip}
\end{equation}
where \(\bm{w}_t\) is the word embedding derived by a one-hot vector, and \(\bm{\bar{v}} = \frac{1}{k}\Sigma_i\bm{v_i}\) denotes the mean pooling of image features. Therefore, we could obtain:
% \vspace{-0.12cm}
\begin{equation}
  \bm{h}_t, \bm{c}_t = LSTM\big(\bm{x}_t,  (\bm{h}_{t-1}, \bm{c}_{t-1})\big)
  % \vspace{-0.12cm}
\end{equation}
In view of the sequentiality output of LSTM, \(\bm{h}_t\) could often be treated as an encoding of sequence order for caption words. Such an encoding could provide its subsequent decoder with precise position information in two aspects. One is what words have been generated by far, and the other is where the decoder is for directing the current relative position of objects. In addition, this position encoding will also be updated at each time step, and meanwhile, it could also guide the decoder to focus on its correlated regions in geometry-aware mode adaptively.

\textbf{ii) Injecting position encodings into the decoder.}

Given a set of refined region features \(\bm{X}^r\) and a group of current position encoding vectors \(\bm{h}_t\), our decoder generates a multi-layer structure sequence \(\bm{\hat{y}}\). As shown in Eq.(\ref{ff}), each decoding layer of our model contains a sub-layer of attention and a sub-layer of feed-forward output. Sometimes, such a layer may look a little redundant. Therefore, in experiments, we consider removing one self-attention sub-layer so as to compress network size and analyzing the influence of removing the attention sub-layer on model performance.
% Different from the original Transformer decoder, each decoder layer in our model consists of an encoder-decoder attention sub-layer and a point-wise feed-forward sub-layer(see Eq. \ref{ff}), respectively.

Furthermore, the group of decoders at the bottom firstly achieves the current position encoding \(\bm{h}_t\) to calculate the \(\bm{Query}\), and uses the region features \(\bm{X}^r\) to calculate the \(\bm{Keys}\) and \(\bm{Values}\). Then, the scaled attention results by dot-product could be obtained from a multi-head pattern. For the back layers of decoder, however, they always adopt the output of previous layers as the \(\bm{Query}\). As a result, at each time step \(t\), through the conditioning on the output of top decoder, the distribution of the $t$-th word could be calculated by:
% \vspace*{-0.4\baselineskip}
\begin{equation}
  p(\bm{y}_t \mid \bm{y}_{1:t-1}) = {softmax} (\bm{W}_p \bm{\mathcal{F}}_{t}^{} + \bm{b}_p)
  % \vspace*{-0.08\baselineskip}
\end{equation}
where \(\bm{W}_p \in \mathbb{R}^{|\Sigma| \times m}\) and \(\bm{b}_p \in \mathbb{R}^{|\Sigma|}\) are also two groups of learnable parameters, and \(\bm{\mathcal{F}}_{t}^{} \) is the final output of designed decoder.

\section{Experiments}
\subsection{Datasets and Metrics}
We evaluate our model on the dataset MS COCO \cite{lin2014microsoft} and dataset Flickr30k \cite{young2014image}. MS COCO is one of the largest datasets for image captioning, consisting of 123,287 images with five captions labeled for each. For validation and off-line comparison, we employ the widely-used ``Karpathy'' split \cite{karpathy2015deep},  which contains 113,287 training images, 5000 for validation and 5000 for testing. Moreover, Flickr30k contains 31,783 images also with five captions for each. For both datasets, we truncate the captions longer than 16 words and convert all sentences into lower case, obtaining an experimental vocabulary with 9,487 words for COCO and 7,000 words for Flickr30k.

Following public convention, there are five classic metrics to be adopted for evaluating the performance of our model. They contain the BLEU \cite{papineni2002bleu}, the METEOR \cite{banerjee2005meteor}, the ROUGE \cite{lin-2004-rouge}, the CIDEr \cite{vedantam2015cider} and the SPICE \cite{spice2016}.

\subsection{Implementation Details}
We employ a pre-trained Faster R-CNN model with a backbone of ResNet-101 \cite{Anderson_2018_CVPR}, to extract 36 features for each image. The dimension of feature vectors is 2048-D. Thereafter we project them into 512-D to reduce memory consumption. The hidden state size of LSTM is set to 1024. In addition, the dimensionality of the input to both the GSR and the self-attention module is set to 512. Moreover, the number of self-attention heads is 8. We set the number of layers to 3 in both the encoder and the decoder. In LSTM, the dropout rate is 0.5, and the dropout rate of all self-attention layers is set to 0.1. In the stage of cross-entropy training, we train our model using an initial learning rate of $(5 \times 10^{-4})$ with a decay rate of 0.8 for every three epochs. In the CIDEr optimizing, we train our model in 30 epochs, by a learning rate of $(2 \times 10^{-5})$ with a decay factor of 0.8 also for every three epochs. All compared models are trained with the Adam optimizer with batch size 50. In testing, we always use the same beam size of five for all.

\subsection{Ablation Experiments}
To quantify the performance of our proposed modules, we design a group of ablative experiments on MS COCO. We use a \emph{vanilla Transformer} (see Figure \ref{fig:transformer}) as our experiment '\emph{base}', also similar to the self-attention network \cite{guo2020normalized}. Its encoders do not consider the geometry information, and decoders are only combined with common sinuous position encodings.

\textbf{i) Effect of Geometry Self-attention}

We first apply the GSR to the '\emph{base}' model to evaluate its effect on encoders. GSR refines raw image representations by injecting explicit geometry relations. From Table \ref{tab:ablative}, we could see that it obtains a huge improvement of CIDEr score (from 109.0 to 115.1) by applying our GSR. This comparison demonstrates that the base model without object geometry relations may be confused by irrelevant regions and misled by wrong information. Admittedly, our GSR furnishes the base model with a sense of '\emph{where}'. It means that the model could look purposefully and thus generate the caption with precise order and geometry-aware words.

\textbf{ii) Effect of Position LSTM}

We further evaluate the impact of the position-LSTM module on captioning performance. We replace the simple sinuous position encoding in \cite{guo2020normalized}, with our LSTM-based ones. From the experimental comparisons in Table \ref{tab:ablative}, we could see that our position-LSTM has raised the CIDEr score of the `base' model by 5.9. Compared with sinuous encoding, our position-LSTM could provide the decoder with an expressive encoding for '\emph{where}' to decode and also guide the decoder to capture correlated semantic information from image regions.
\begin{table}[tb]
  % \vspace{-0.2cm}
  % \footnotesize
  \large
  \begin{center}
    \caption{Ablation experiment comparisons. The results are reported after cross-entropy loss stage on the dataset ``Karpathy'' test split \cite{karpathy2015deep}.}
    \label{tab:ablative}
    \resizebox{0.9\textwidth}{!}{
      \begin{tabular}{l | c c c c c c}
        \toprule
        % Model & B@1 & B@4 & M & R & C & S\\
        Model              & BLEU-1        & BLEU-4        & METEOR        & ROUGE         & CIDEr          & SPICE         \\
        \midrule
        Base               & 75.0          & 32.8          & 27.3          & 55.5          & 109.0          & 20.6          \\
        % \hline
        Base+GSR           & 76.9          & 35.6          & 28.1          & 57.0          & 115.1          & 21.4          \\
        % \hline
        Base+position-LSTM & 76.5          & 34.5          & 28.0          & 56.8          & 114.9          & 21.3          \\
        \hline
        Full: GAT          & \textbf{77.5} & \textbf{37.8} & \textbf{28.5} & \textbf{57.6} & \textbf{119.8} & \textbf{21.8} \\
        \bottomrule
      \end{tabular}}
  \end{center}
  % \vspace{-0.8cm}
\end{table}

\begin{table}[tb]
  % \vspace{-0.2cm}
  % \footnotesize
\LARGE
  \begin{center}
    \caption{The comparison results of different component combination strategies, including geometry Queries and Keys, and the GLU module in the Transformer structure.}
    \label{tab:ablative_more}
    \resizebox{0.9\textwidth}{!}{
      \begin{tabular}{l | c c c c c c c}
        \toprule
        Modules & Strategy     & BLEU-1        & BLEU-4        & METEOR        & ROUGE         & CIDEr          & SPICE         \\
        \midrule
        \multirow{2}*{For Geometry Q \& K} & \emph{add}.          & 76.0          & 35.1          & 27.2          & 56.0          & 116.4          & 20.7          \\
        ~                            & \emph{concat}.       & \textbf{77.5} & \textbf{37.8} & \textbf{28.5} & \textbf{57.6} & \textbf{119.8} & \textbf{21.8} \\
        \midrule
        \multirow{3}*{For GLU}           & without GLU       & 76.7          & 35.2          & 27.6          & 56.5          & 114.9          & 21.2          \\
        ~                            & GLU(enc.)     & \textbf{77.7} & \textbf{37.8} & \textbf{28.4} & \textbf{57.4} & \textbf{119.4} & \textbf{21.8} \\
        ~                            & GLU(enc. and dec.) & 77.4          & \textbf{37.8} & 28.2          & 57.1          & 118.9          & 21.5          \\
        \bottomrule
      \end{tabular}}
  \end{center}
  % \vspace{-0.8cm}
\end{table}

\textbf{iii) Geometry Queries and Keys}

To verify the efficiency of concatenating geometrical queries and keys with appearance queries and keys, we investigate different straggles to combine them. Mainly, we directly compare two operations, including '\emph{add}' (adding) and '\emph{concat}' (concatenating). Table \ref{tab:ablative_more} shows the comparison results of these two operations. We could figure out that the performance promotion by concatenating appearance queries and keys with geometry queries and keys is better than by adding them, although they both could surpass the base model in terms of most metrics. For example, the BLEU-1 score of the base model is only 75.0, while it could rise to 76.0 and 77.5 by '\emph{add}' and '\emph{concat}', respectively.

\textbf{iv) Gated Linear Unit.}
In our architecture, GLUs play an important role in refining the output of original self-attention layers. Therefore we also have tested its effect on both encoders and decoders. Table \ref{tab:ablative_more} lists the comparison results. We could see that GLUs could always achieve a good performance promotion on the base model. However, we could also see that the GLU works best when it cooperates alone with encoders. If GLUs are assembled with encoders and decoders at the same time, the model could even get worse performance. For example, in Table \ref{tab:ablative_more}, the SPICE by the base model without GLUs is 21.2, and it gets to 21.8 when GLUs are integrated only with encoders. However, the SPICE drops to 21.5 when GLUs are combined with all encoders and decoders. This finding is exactly consistent with \cite{huang2019attention}. It means that stacking too many GLUs on a group of position-LSTM decoder layers could damage the gradient of network training, depressing the capacity of self-attention layers.

\subsection{Comparison with State-of-the-Art Models}
We have also compared our GAT model with several state-of-the-art approaches of recent years. On dataset MS COCO, these comparison methods mainly include the SCST \cite{rennie2017self} optimizing evaluation metrics directly, the Up-Down model \cite{Anderson_2018_CVPR} with two-layer LSTM structure for extracting bottom-up features, and the ORT \cite{herdade2019image} employing a Transformer-like model with an object relation module. Besides, we also compare our model with the RF-Net \cite{jiang2018recurrent} which employs features by multiple CNNs, the GCN-LSTM \cite{yao2018exploring} exploiting pairwise relations by Graph Convolutional Network, the GAE \cite{yang2019auto} via auto-encoding scene graphs, the AoANet \cite{huang2019attention} refining self-attention results also by GLUs. On Flickr30k, the models compared with ours include the Soft-Attention \& Hard-Attention \cite{xu2015show}, the Deep VS \cite{karpathy2015deep}, the NIC \cite{vinyals2015show}, the m-RNN \cite{mao2014deep}, the adaptive model \cite{Lu_2017_CVPR}, the SEM architecture \cite{CAI202031} and the DA framework \cite{gao2019deliberate}.

\linespread{2}
\begin{table*}[htb]
\LARGE
  % \vspace{-0.5cm}
  % \footnotesize
  % \scriptsize
  \begin{center}
    \caption{The off-line comparisons with some state-of-the-art methods on the '\emph{Karpathy}' test splits of MS COCO. The higher, the better for all data in columns. The visual features of compared models are extracted by  a same structure with the Faster RCNN \cite{Anderson_2018_CVPR}. Our model has reduced the dimensionality of embedding from $2048$ to $512$, due to the limitation of main memory capability.}
    \vspace{0.12cm}
    \label{tab:offline}
    \resizebox{1\textwidth}{!}{
      \begin{tabular}{l | c c c c c c | c c c c c c}
        \toprule
        % \hline
        Model                                       & \multicolumn{6}{c}{\textbf{Cross-Entropy Loss}} & \multicolumn{6}{c}{\textbf{CIDEr-D Optimization}}                                                                                                                                                                    \\
        \hline
                                              & BLEU-1                                          & BLEU-4                                            & METEOR        & ROUGE          & CIDEr          & SPICE         & BLEU-1        & BLEU-4        & METEOR        & ROUGE         & CIDEr          & SPICE         \\
        \midrule
        % \hline
                                                    & \multicolumn{12}{c}{\textbf{Single Model}}                                                                                                                                                                                                                             \\
        \midrule
        SCST \cite{rennie2017self}                  & -                                               & 30.0                                              & 25.9          & 53.4           & 99.4           & -             & -             & 34.2          & 26.7          & 55.7          & 114.0          & -             \\
        Up-Down \cite{Anderson_2018_CVPR}           & 77.2                                            & 36.2                                              & 27.0          & 56.4           & 113.5          & 20.3          & 79.8          & 36.3          & 27.7          & 56.9          & 120.1          & 21.4          \\
        RFNet \cite{jiang2018recurrent}             & 76.4                                            & 35.8                                              & 27.4          & 56.8           & 112.5          & 20.5          & 79.1          & 36.5          & 27.7          & 57.3          & 121.9          & 21.2          \\
        GCN-LSTM \cite{yao2018exploring}            & 77.3                                            & 36.8                                              & 27.9          & 57.0           & 116.3          & 20.9          & 80.5          & 38.2          & 28.5          & 58.3          & 127.6          & 22.0          \\
        SGAE \cite{yang2019auto}                    & -                                               & -                                                 & -             & -              & -              & -             & \textbf{80.8} & 38.4          & 28.4          & 58.6          & 127.8          & 22.1          \\
        ORT \cite{herdade2019image}                 & -                                               & -                                                 & -             & -              & -              & -             & 80.5          & 38.6          & 28.7          & 58.4          & 128.3          & 22.6          \\
        AoANet \cite{huang2019attention}            & 77.4                                            & 37.2                                              & 28.4          & 57.5           & \textbf{119.8} & 21.3          & 80.2          & 38.9          & 29.0          & 58.8          & 129.8          & 22.4          \\
        \hline
        GAT (ours)                                  & \textbf{77.5}                                   & \textbf{37.8}                                     & \textbf{28.5} & \textbf{57.6}  & \textbf{119.8} & \textbf{21.8} & \textbf{80.8} & \textbf{39.7} & \textbf{29.1} & \textbf{59.0} & \textbf{130.5} & \textbf{22.9} \\
        \midrule
                                                    & \multicolumn{12}{c}{\textbf{Ensemble/Fusion}}                                                                                                                                                                                                                          \\
        \midrule
        SCST~\cite{rennie2017self}$^{\Sigma}$       & -                                               & 32.8                                              & 26.7          & 55.1           & 106.5          & -             & -             & 35.4          & 27.1          & 56.6          & 117.5          & -             \\
        RFNet~\cite{jiang2018recurrent}$^{\Sigma}$  & 77.4                                            & 37.0                                              & 27.9          & 57.3           & 116.3          & 20.8          & 80.4          & 37.9          & 28.3          & 58.3          & 125.7          & 21.7          \\
        GCN-LSTM~\cite{yao2018exploring}$^{\Sigma}$ & 77.4                                            & 37.1                                              & 28.1          & 57.2~          & 117.1          & 21.1          & 80.9          & 38.3          & 28.6          & 58.5          & 128.7          & 22.1          \\
        SGAE~\cite{yang2019auto}$^{\Sigma}$         & -                                               & -                                                 & -             & -              & -              & -             & 81.0          & 39.0          & 28.4          & 58.9          & 129.1          & 22.2          \\
        AoANet \cite{huang2019attention}$^{\Sigma}$ & 78.7                                            & 38.1                                              & 28.5          & 58.2           & 122.7          & 21.7          & \textbf{81.6} & 40.2          & 29.3          & 59.4          & 132.0          & 22.8          \\
        \hline
        GAT (Ours)$^{\Sigma}$                       & \textbf{79.0}                                   & \textbf{38.8}                                     & \textbf{28.7} & \textbf{58.9 } & \textbf{123.7} & \textbf{22.1} & \textbf{81.6} & \textbf{40.7} & \textbf{29.4} & \textbf{59.6} & \textbf{133.4} & \textbf{23.2} \\
        \bottomrule
      \end{tabular}}
  \end{center}
  % \vspace{-0.8cm}
\end{table*}

\textbf{i) Off-line Evaluation}

We also evaluate our model on the '\emph{Karpathy}' test split of MS COCO, as in \cite{karpathy2015deep}. All models are firstly trained by cross-entropy loss and then optimized for the CIDEr score. For a fair comparison, the visual features fed to all models are directly extracted by a Faster R-CNN of the same structure. The top half of Table \ref{tab:offline} ($i.e.$, Single Model) shows the performance comparisons of all models. By comparisons, it could be seen that our GAT outperforms all other methods on almost all metrics in terms of cross-entropy loss and CIDEr Optimization. Compared with the ORT \cite{herdade2019image}, on {CIDEr-D} optimization, our GAT has obtained a BLEU-4 rise of $1.1\%$, METEOR rise of 0.4\%, ROUGE rise of 0.6\%, CIDEr rise of 2.2\%, and SPICE rise of 0.3\% by a single model. These rise values look obvious. Especially on BLEU-4, it achieves an increase from 38.6 to 39.7, and on CIDEr, it rises from 128.3 to 130.5. Similarly, our GAT also outperforms the AOANet \cite{huang2019attention}, in terms of almost all metrics except the CIDEr of Cross-Entropy.

Moreover, in the down half of Table \ref{tab:offline} ($i.e.$, Ensemble/Fusion), it shows the comparisons of 6 different captioning models. Every model is run four times respectively under four different initialization conditions, and then only the mean of four outputs is used for the comparisons of captioning models. We could also see that our GAT could surpass all other models on almost all metrics while optimized by Cross-Entropy Loss. Even on the metrics of CIDEr-D optimization, our BLEU-1 could almost keep the same as the AOANet \cite{huang2019attention}, the best of all other models. On the rest metrics of CIDEr-D optimization, our GAT could always outperform all other models. For example, on the CIDEr of Cross-Entropy loss, our model achieves a result of 123.7, just 1.0 larger than the AOANet \cite{huang2019attention}. Moreover, the CIDEr score of CIDEr-D optimization of our model rises to 133.4, while the one of AOANet \cite{huang2019attention} is only 123.0, about 1.4 less than ours.

\begin{table*}[htb]
\footnotesize
  \begin{center}
    \caption{The off-line comparison results with state-of-the-art methods on dataset Flickr30k. The higher, the better for the data  in all data columns.}
    \vspace{0.15cm}
    \label{tab:flickr30k}
    \resizebox{1 \textwidth}{!}{
      \begin{tabular}{l | c c c c c c }
        \toprule
                            & BLEU-1        & BLEU-2        & BLEU-3        & BLEU-4        & METEOR        & CIDEr         \\
        \midrule
        Deep VS \cite{karpathy2015deep}   & 57.3          & 36.9          & 24.0          & 15.7          & 15.3          & 24.7          \\
        Soft-Attention \cite{xu2015show}  & 66.7          & 43.4          & 28.8          & 19.1          & 18.5          & -             \\
        Hard-Attention \cite{xu2015show}  & 66.9          & 43.9          & 29.6          & 19.9          & 18.5          & -             \\
        Google NIC \cite{vinyals2015show} & 66.4          & 42.3          & 27.7          & 18.3          & -             & -             \\
        m-RNN \cite{mao2014deep}          & 60.0          & 41.0          & 28.0          & 19.0          & -             & -             \\
        Adaptive \cite{Lu_2017_CVPR}      & 67.6          & 49.4          & 35.4          & 25.1          & 20.4          & 53.1          \\
        SEM \cite{CAI202031}              & 73.1          & 55.1          & 40.1          & 29.0          & 22.0          & 66.8          \\
        DA \cite{gao2019deliberate}       & 73.8          & 55.1          & 40.3          & 29.4          & 23.0          & 66.6          \\
        \hline
        GAT (ours)                        & \textbf{74.4} & \textbf{56.7} & \textbf{41.8} & \textbf{30.8} & \textbf{23.4} & \textbf{68.0} \\
        \bottomrule
      \end{tabular}}
  \end{center}
  % \vspace{-0.8cm}
\end{table*}

In Table \ref{tab:flickr30k}, we compare the performance of our GAT with state-of-the-art models on dataset Flickr30k. It could be seen that our model outperforms all the compared methods by a large margin. For example, compared with Soft-Attention, which also employs an attention mechanism \cite{xu2015show}, our GAT obtains an improvement of 7.5 on BLEU-1, 10.9 on BLEU-4 and 4.9 on METEOR. Even compared with DA model\cite{gao2019deliberate}, the best of the other eight models, our GAT still could often achieve a higher score than DA on all metrics. For example, our GAT could get an increase of about 0.6, 1.4 and 1.4 on BLEU-1, BLEU-4 and CIDEr, respectively. This comparison further demonstrates the superiority of our GAT method.

% \footnote{\scriptsize\url{https://competitions.codalab.org/competitions/3221\#results}}
\textbf{ii) On-line Evaluations}

\linespread{2}
\begin{table*}[htb]
\LARGE
  % \scriptsize
  \begin{center}
    \caption{The leaderboard of various methods on the online test server specially for COCO. Our method, GAT, went into the top 10 on the leaderboard when submitted online, on June 5th, 2021. The higher, the better for all listed data.}
     \vspace{0.16cm}
    \label{tab:online}
    \resizebox{1\textwidth}{!}{
      \begin{tabular}{l| c c c c c c c c c c c c c c}
        \toprule
        Model                             & \multicolumn{2}{c}{BLEU-1} & \multicolumn{2}{c}{BLEU-2} & \multicolumn{2}{c}{BLEU-3} & \multicolumn{2}{c}{BLEU-4} & \multicolumn{2}{c}{METEOR} & \multicolumn{2}{c}{ROUGE} & \multicolumn{2}{c}{CIDEr}                                                                                                                   \\
        \hline
                                    & c5                         & c40                        & c5                         & c40                        & c5                         & c40                       & c5                        & c40           & c5            & c40           & c5            & c40           & c5             & c40            \\
        \midrule
        SCST~\cite{rennie2017self}        & 78.1                       & 93.7                       & 61.9                       & 86.0                       & 47.0                       & 75.9                      & 35.2                      & 64.5          & 27.0          & 35.5          & 56.3          & 70.7          & 114.7          & 116.0          \\
        Up-Down~\cite{Anderson_2018_CVPR} & 80.2                       & 95.2                       & 64.1                       & 88.8                       & 49.1                       & 79.4                      & 36.9                      & 68.5          & 27.6          & 36.7          & 57.1          & 72.4          & 117.9          & 120.5          \\
        RFNet~\cite{jiang2018recurrent}   & 80.4                       & 95.0                       & 64.9                       & 89.3                       & 50.1                       & 80.1                      & 38.0                      & 69.2          & 28.2          & 37.2          & 58.2          & 73.1          & 122.9          & 125.1          \\
        GCN-LSTM~\cite{yao2018exploring}  & -                          & -                          & 65.5                       & 89.3                       & 50.8                       & 80.3                      & 38.7                      & 69.7          & 28.5          & 37.6          & 58.5          & 73.4          & 125.3          & 126.5          \\
        SGAE~\cite{yang2019auto}          & 81.0                       & \textbf{95.3}              & 65.6                       & 89.5                       & 50.7                       & 80.4                      & 38.5                      & 69.7          & 28.2          & 37.2          & 58.6          & 73.6          & 123.8          & 126.5          \\
        AoANet~\cite{huang2019attention}  & 81.0                       & 95.0                       & 65.8                       & 89.6                       & 51.4                       & 81.3                      & 39.4                      & 71.2          & \textbf{29.1} & \textbf{38.5} & 58.9          & \textbf{74.5} & 126.9          & 129.6          \\
        NG-SAN~\cite{guo2020normalized}   & 80.8                       & 95.0                       & 65.4                       & 89.3                       & 50.8                       & 80.6                      & 38.8                      & 70.2          & 29.0          & 38.4          & 58.7          & 74.0          & 126.3          & 128.6          \\
        \hline
        GAT (Ours)                        & \textbf{81.1}              & 95.1                       & \textbf{66.1}              & \textbf{89.7}              & \textbf{51.8}              & \textbf{81.5}             & \textbf{39.9}             & \textbf{71.4} & \textbf{29.1} & 38.4          & \textbf{59.1} & 74.4          & \textbf{127.8} & \textbf{129.8} \\
        \bottomrule
      \end{tabular}}
  \end{center}
\end{table*}

For a fairer comparison and getting to know the true position of our GAT in the image captioning task, we also submitted it to the official MS COCO test server. The last line in Table \ref{tab:online} lists the test results of our GAT model. In this test, an ensemble of 4 different outputs of our model trained by the `\emph{Karpathy}' split is used for comparing with others. From Table \ref{tab:online}, we could see that our GAT obtains the best captioning performance on each c5 of almost all metrics, and it's only a bit lower than others on the c40 of three metrics. For example, it's 127.8 on CIDEr(c5), almost 0.9 higher than  AoANet \cite{huang2019attention}, the best of the others. Even on BLEU-1(C40) and ROUGE(c40), our GAT is only paltry 0.2 lower than the SGAE \cite{yang2019auto} and 0.1 lower than the AoANet \cite{huang2019attention}. Therefore, on the whole, our GAT model could be synthetically deemed to be more superior than other compared ones.

\subsection{Caption Text Comparisons}
\begin{figure*}[htb]
  \begin{center}
    \includegraphics[width=1.0\textwidth]{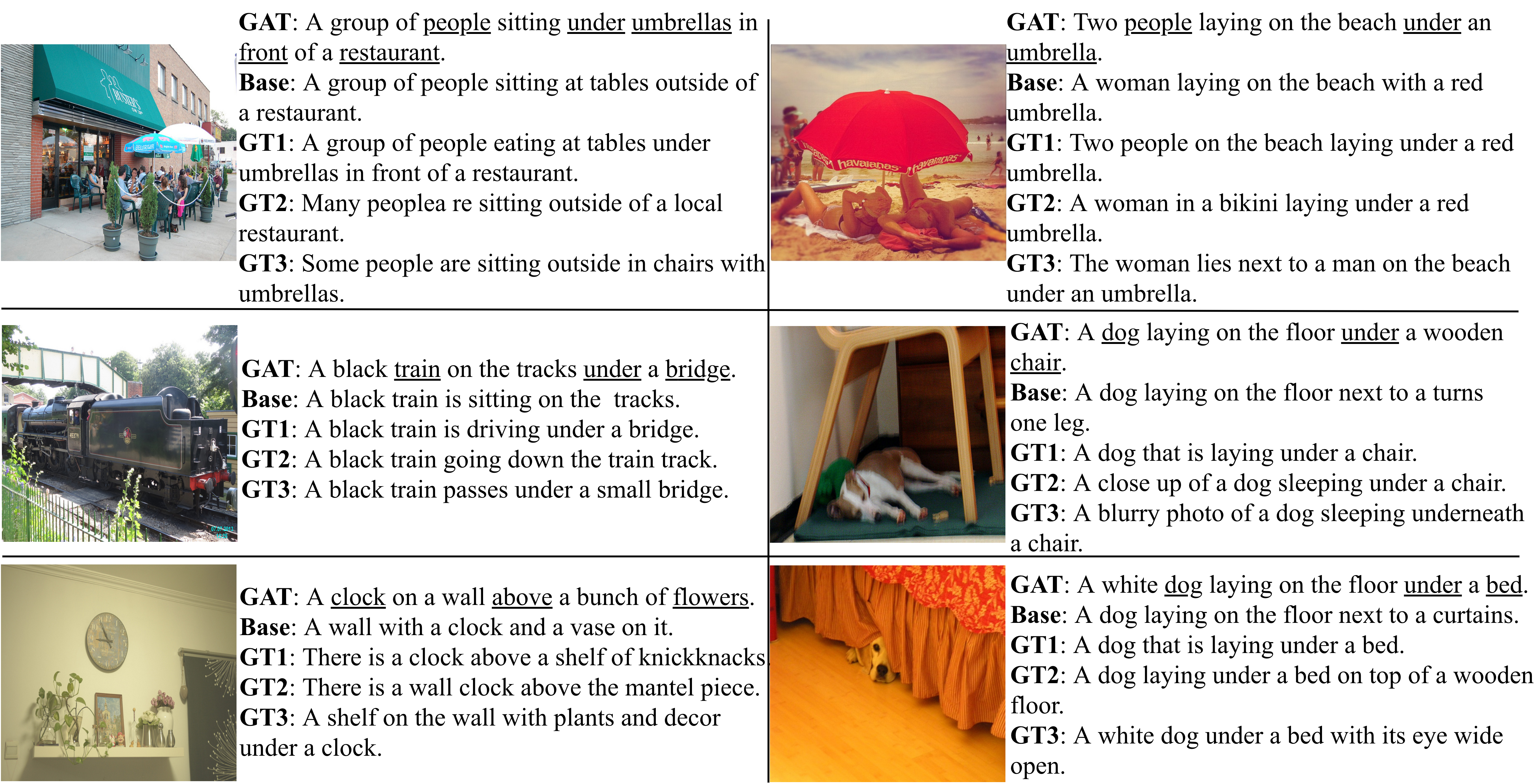}
  \end{center}
  % \vspace{-0.65cm}
  \caption{The output comparison of captions generated respectively by our GAT and a base model, as well as three manual description captions of ground truth, GT1$\sim$GT3.}
  \label{fig:location}
  % \vspace{-0.6cm}
\end{figure*}

To qualitatively illustrate the superiority of our method, we present six images with their captions generated respectively by our GAT and the \emph{Vanilla Transformer} model ($i.e.$, base), as well as three manual captions of ground truth ($i.e.$, GT1$\sim$ GT3) in Figure \ref{fig:location}. It can be seen that GAT could generate an accurate image caption with complete semantic structure and clear position relations. For example, the caption by GAT could geometrically describe that ``people'' are ``\textit{under} umbrellas'' and ``in \textit{front} of store'' in the top-left sub-image of Figure \ref{fig:location}, and ``dog laying \textit{under} the chair'' rather than ``laying \emph{on} the chair' in the middle-right sub-image of Figure \ref{fig:location}.

The novel characteristic that our GAT is able to accurately describe the geometry \& position relations mainly owes to the GSR module, which can explicitly incorporate the spatial correlations of image regions into object feature representations. Moreover, the position-LSTM could also keep reminding the decoder, at every decoding step, which visual object should be attended to. On the whole, our GAT could often express almost all exact objects contained in a given image, such as ``\emph{people}'', ``\emph{umbrellas}'' and ``\emph{restaurant}'', etc. in the top-left sub-image of \ref{fig:location}. Furthermore, GAT could also capture the geometry relations between any two detected objects and then decide whether to specify the correlations of detected objects, benefited by which it could often generate the highly-reliable image captions with spatial-aware semantic.

% Finally, we further investigate the caption generation process at each decoding step. Results are shown in Figure . It can be seen that our GAT correctly attends to image regions correlated with caption words and is capable of reasoning the relative position among different objects.

\section{Conclusions}
In this paper, we propose the Geometry Attention Transformer (GAT), an extension scheme to the well-known Transformer for image captioning in recent years. It is able to explicitly refine image representations by incorporating the geometry features of visual objects into region encodings. Moreover, the position-LSTM in decoder layers could often fulfill a precise encoding for the word order of caption texts. The ablation experiments on the base model show that the GSR for capturing geometry features and the position-LSTM for injecting position encodings could be effective. Each of them, if cooperating with a base model, could promote image captioning performance. In addition, the experimental comparisons (off-line $\&$ on-line) also show that our GAT framework could often outperform state-of-the-art ones on the datasets MS COCO and Flickr30k, respectively.

% caption models indeed benefit from relative geometry information and our approach achieves state-of-the-art performance on datasets, MS COCO and Flickr30K. Moreover, GAT can be easily applied to any other Transformer-like architecture.

\section*{Acknowledgement}
This work is supported by the National Natural Science Foundation of China (NSFC) under Grant No. 61972072. We would like to thank the Editors and Reviewers for possible revision suggestions.

\scriptsize
\bibliography{mybibfile}

\end{document}